\documentclass[10pt,twocolumn]{ICCAS}

\usepackage{graphics}           
\usepackage{times}              
\usepackage{amsmath}            
\usepackage{amssymb}            
\usepackage{graphicx}
\usepackage{algorithm}
\usepackage[noend]{algpseudocode}
\usepackage{booktabs}
\usepackage{makecell}
\usepackage{color}
\usepackage{multirow}
\usepackage{subcaption}
\usepackage{rotating}
\usepackage{cite}
\definecolor{instructioncolor}{rgb}{.5,.5,.5}

\makeatletter
\newcommand\xref[1]{\mbox{\ref{#1}\hskip 1em\hskip-\fontdimen2\font\relax}}
\makeatother
\def\secref#1{Sec.~\xref{#1}}

\def\eqref#1{(\ref{#1})}

\newcommand{\rom}[1]{\uppercase\expandafter{\romannumeral #1\relax}}

\makeatletter
\usepackage{xspace}
\DeclareRobustCommand\onedot{\futurelet\@let@token\@onedot}
\def\@onedot{\ifx\@let@token.\else.\null\fi\xspace}

\makeatother

\usepackage{array}
\newcolumntype{L}[1]{>{\raggedright\let\newline\\\arraybackslash\hspace{0pt}}m{#1}}
\newcolumntype{C}[1]{>{\centering\let\newline\\\arraybackslash\hspace{0pt}}m{#1}}
\newcolumntype{R}[1]{>{\raggedleft\let\newline\\\arraybackslash\hspace{0pt}}m{#1}}

\renewcommand{\b}[1]{\mbox{\boldmath$#1$}}

\newcommand{\dR}{\b R}
\newcommand{\dS}{\b S}

    \newcommand\dLidarFrame{\mathit{L}}
    \newcommand\dCameraFrame{\mathit{C}}
    \newcommand\dRadarFrame{\mathit{R}}

        \newcommand\dRot{\mathbf{R}}

        \newcommand\dRotFromFrameLidarToRadar{\dRot_{\dLidarFrame\dRadarFrame}}

        \newcommand\dRotPert{\boldsymbol{\theta}}

        \newcommand\dRotPertFromFrameLidarToRadar{\dRotPert_{\dLidarFrame\dRadarFrame}}

        \newcommand\dTls{\mathbf{t}}

        \newcommand\dTlsFromFrameLidarToRadar{\dTls_{\dLidarFrame\dRadarFrame}}

        \newcommand\dTlsPert{\tilde{\dTls}}

        \newcommand\dTlsPertFromFrameLidarToRadar{\dTlsPert_{\dLidarFrame\dRadarFrame}}

        \newcommand\dPt{\mathbf{p}}

        \newcommand\dRadarPt{\dPt_{\dRadarFrame}}

        \newcommand\dRadarPtInRadarFrame{{}^{\dRadarFrame} \dRadarPt}
        \newcommand\dRadarPtInLidarFrame{{}^{\dLidarFrame} \dRadarPt}

        \newcommand\dJthRadarPt{\dPt_{\dRadarFrame_j}}
        \newcommand\dLthRadarPt{\dPt_{\dRadarFrame_l}}

        \newcommand\dKthLidarPt{\dPt_{\dLidarFrame_k}}
        \newcommand\dLthLidarPt{\dPt_{\dLidarFrame_l}}

        \newcommand\dJthRadarPtInCameraFrame{{}^{\dCameraFrame} \dJthRadarPt}

        \newcommand\dKthLidarPtInCameraFrame{{}^{\dCameraFrame} \dKthLidarPt}

        \newcommand\dPtNom{\hat{\dPt}}

        \newcommand\dJthRadarPtNom{\dPtNom_{\dRadarFrame_j}}
        \newcommand\dLthRadarPtNom{\dPtNom_{\dRadarFrame_l}}

        \newcommand\dKthLidarPtNom{\dPtNom_{\dLidarFrame_k}}
        \newcommand\dLthLidarPtNom{\dPtNom_{\dLidarFrame_l}}

        \newcommand\dRadarPtNomInLidarFrame{{}^{\dLidarFrame} \dPtNom_{\dRadarFrame}}

        \newcommand\dJthRadarPtNomInCameraFrame{{}^{\dCameraFrame} \dJthRadarPtNom}
        \newcommand\dLthRadarPtNomInCameraFrame{{}^{\dCameraFrame} \dLthRadarPtNom}

        \newcommand\dKthLidarPtNomInCameraFrame{{}^{\dCameraFrame} \dKthLidarPtNom}
        \newcommand\dLthLidarPtNomInCameraFrame{{}^{\dCameraFrame} \dLthLidarPtNom}

        \newcommand\dRadarPtNoiseInLidarFrame{{}^{\dLidarFrame} \mathbf{n}_{\dRadarFrame}}

\def\dSE3{\mathrm{SE}(3)}
\def\dSO3{\mathrm{SO}(3)}
\def\dS2{\mathbb{S}^2}
\def\dR3{\mathbb{R}^3}
\def\dR6{\mathbb{R}^6}

\newcommand\dBigExp{\mathrm{Exp}}

\makeatletter
\def\fix@mathalph#1{%
  \expandafter\let\csname decl@#1\expandafter\endcsname\csname #1 \endcsname
  \expandafter\DeclareRobustCommand\csname #1\endcsname[1]{{\csname decl@#1\endcsname ##1}}}
\fix@mathalph{mathcal}
\fix@mathalph{mathbf}
\fix@mathalph{mathrm}
\fix@mathalph{mathit}
\fix@mathalph{mathsf}
\fix@mathalph{mathtt}
\fix@mathalph{mathbb}
\fix@mathalph{mathfrak}
\makeatother

\makeatletter
\let\ICCASnormalsize\normalsize
\renewcommand\normalsize{%
  \ICCASnormalsize
  \abovedisplayskip 7pt plus 2pt minus 4pt\relax
  \belowdisplayskip 7pt plus 2pt minus 4pt\relax
  \abovedisplayshortskip 0pt plus 2pt\relax
  \belowdisplayshortskip 3pt plus 2pt minus 2pt\relax}
\makeatother

\begin{document}

\title{Online Target-less Radar-LiDAR-Camera Extrinsic Calibration\\ via Joint Optimization}

\author{Gunhee Shin${}^{1}$, Yunsoo Kim${}^{1}$, Chanhyuk Lee${}^{1}$, Wanhee Kim${}^{1}$, Minwoo Lee${}^{2}$,\\ Sungwoo Han${}^{3}$, Jeongwoo Woo${}^{3}$, Hyuntai Chin${}^{3}$, Minha Park${}^{3}$, and Hyun Myung${}^{1*}$}

\affils{ ${}^{1}$School of Electrical Engineering, KAIST (Korea Advanced Institute of Science and Technology), \\
Daejeon, 34141, Korea (\{gunhee\_shin, dbstn1121, bright2001, gml78905, hmyung\}@kaist.ac.kr) \\
${}^{2}$Kookmin University, Seoul, 02707, Korea (mw0508@kookmin.ac.kr) \\
${}^{3}$LG Innotek, Seoul, 07796, Korea (\{sungwoo1.han, bluesky0327, hyuntai.chin, minha.park\}@lginnotek.com) \\
{\small${}^{*}$ Corresponding author}}

\abstract{
  Fusing radar, LiDAR, and camera enables robust perception in diverse and adverse conditions, but the fusion performance critically depends on accurate extrinsic calibration among the three sensors.
  In this paper, we address the problem of online target-less extrinsic calibration for the radar-LiDAR-camera system.
  Existing target-less methods are mostly designed for a single sensor pair, and composing the pairwise results does not guarantee consistency across the three sensors.
  Moreover, the sparse and noisy radar measurements make the radar-involving pairs unreliable.
  To tackle these challenges, we propose a joint calibration framework that constructs residuals for each sensor pair and optimizes the extrinsics of all pairs together to minimize the overall residual.
  Furthermore, we introduce an adaptive radar noise filter that rejects spurious radar returns using a range-dependent margin, and a correspondence accumulation strategy that aggregates sparse radar correspondences over frames.
  We validate our method on an in-house radar-LiDAR-camera dataset covering diverse urban environments, where it reduces calibration errors across all sensor pairs over a state-of-the-art camera-LiDAR baseline.
}


\maketitle

\section{Introduction}

Multiple modalities have been actively explored in recent robotics research, especially for perception and localization tasks.
Among various modalities, camera, LiDAR, and radar are commonly used due to their complementary characteristics.
Cameras provide rich texture and color information but are sensitive to lighting and weather conditions~\cite{zhao2025Resilient, lee2022ral-vivid}.
LiDARs offer accurate 3D geometric information even in low-light conditions, but may degrade in geometrically degenerate environments such as long tunnels or featureless open roads~\cite{xu2022FASTLIO2, shin2026SE3LIO, lim2023adalio}.
Radars, in contrast, remain robust under adverse weather conditions such as fog and rain where other sensors tend to degrade, and can directly estimate velocity, although they typically produce noisy measurements and have lower resolution~\cite{heng2020Automatic, cheng2023Online, lim2023icra}.

By fusing data from these sensors, systems can achieve more robust and reliable performance in diverse environments~\cite{zhao2025Resilient, xu2022FASTLIO2, shin2026SE3LIO}.
However, to ensure fusion performance, several considerations must be carefully addressed, including appropriate sensor placement, extrinsic calibration, and temporal synchronization~\cite{zhao2025Resilient, park2025Huber}.
When these considerations are not properly handled, the fused system can even perform worse than a single-sensor system.
Among these considerations, extrinsic calibration, which estimates the 6-DoF relative transformation between sensors so that their measurements share a common coordinate frame, is particularly critical~\cite{beltran2022Automatic, domhof2019Extrinsic}.

Extrinsic calibration methods can be widely categorized into target-based and target-less approaches.
Target-based approaches rely on hand-crafted calibration targets, such as checkerboards or AprilTags, with known geometry.
They yield reliable correspondences and high accuracy, but require dedicated targets and a controlled environment, which limits their flexibility~\cite{beltran2022Automatic}.
On the other hand, target-less approaches exploit natural features in the environment, such as planes or edges.
They are flexible and applicable to diverse environments, but tend to be less accurate and more dependent on the surrounding environment, which limits their reliability~\cite{yuan2021PixelLevel, song2024Galibr, koide2023General}.

Owing to this flexibility, target-less calibration is widely used for online re-calibration in real-world deployments, especially for autonomous vehicles and field robots traversing unstructured environments, where aggressive motion and external impacts can shift the sensor alignment.
However, most existing methods focus on individual sensor pairs, such as camera-LiDAR~\cite{yuan2021PixelLevel, song2024Galibr, koide2023General, lv2021LCCNeta, cattaneo2025CMRNext, ou2025Iterative}, LiDAR-radar~\cite{heng2020Automatic}, camera-radar~\cite{cheng2023Online, scholler2019Targetless}, and camera-GNSS/INS~\cite{choe2025TwoStage}.
When operating a radar-LiDAR-camera system with such pairwise tools, calibrating each pair independently increases system complexity and yields mutually inconsistent extrinsic estimates.
Joint calibration of all three sensors is therefore needed, yet few works have addressed this problem.
To the best of our knowledge, the only existing approach~\cite{domhof2019Extrinsic} still requires a specially designed target.

In this paper, we address online target-less extrinsic calibration for radar, LiDAR, and camera sensors.
We propose a joint calibration framework that constructs residuals for each sensor pair and optimizes the extrinsics of all pairs together to minimize the overall residual.
In addition, we introduce two complementary mechanisms to handle the sparsity and noise of radar measurements that can destabilize the joint optimization: an adaptive noise filter that rejects spurious radar returns, and a correspondence accumulation strategy that aggregates radar correspondences across frames.
We validate our method on an in-house radar-LiDAR-camera dataset covering diverse urban environments, where it reduces calibration errors across all sensor pairs over a state-of-the-art camera-LiDAR baseline, with the radar-camera rotation error reduced by approximately 65\%.

\section{Related Works}

Target-based extrinsic calibration uses artificial targets with known geometry, such as checkerboards or custom boards with radar reflectors.
Each sensor detects the target through its own features, and the extrinsics are recovered by aligning these detections across sensors~\cite{beltran2022Automatic}.
This yields well-defined correspondences and high accuracy, but the controlled setup limits these methods to offline calibration before deployment.

\subsection{Pairwise Target-less Extrinsic Calibration}

Camera-LiDAR calibration is the most actively studied pairwise problem.
Conventional methods extract geometric features such as planes or edges from the scene and align them across modalities by minimizing photometric or geometric residuals~\cite{yuan2021PixelLevel, koide2023General}.
Galibr~\cite{song2024Galibr} instead used the ground plane~\cite{lim2021ral-patch} to estimate the extrinsics without relying on explicit feature correspondences, reducing the dependence on local feature quality.
While effective in feature-rich scenes, these methods are sensitive to the density of features and require careful parameter tuning across environments.

Learning-based methods estimate the extrinsics directly from raw images and point clouds using neural networks.
LCCNet~\cite{lv2021LCCNeta} introduced an end-to-end network with iterative pose refinement, and CMRNext~\cite{cattaneo2025CMRNext} and Ou et al.~\cite{ou2025Iterative} improved generalization through cross-modal feature matching and iterative pose updates.
While these methods generalize well across scenes, they require large multi-modal datasets for training and do not naturally extend to other sensor pairs such as those involving radar.

Calibration involving radar is less explored, mainly because the sparsity and noise of radar measurements make direct correspondences difficult~\cite{lim2023icra}.
Heng~\cite{heng2020Automatic} calibrated LiDAR and radar by matching their ego-motion estimates, avoiding the need for per-frame point association.
For camera-radar calibration, Sch\"oller et al.~\cite{scholler2019Targetless} matched radar reflectivity patterns to image edges, and Cheng et al.~\cite{cheng2023Online} estimated the extrinsics online using vehicle motion and scene-level priors.
However, these methods focus on a single radar-related pair and cannot be directly extended to a three-sensor system that includes camera, LiDAR, and radar.

\subsection{Joint Multi-sensor Extrinsic Calibration}

Joint calibration of camera, LiDAR, and radar has received little attention.
Domhof et al.~\cite{domhof2019Extrinsic} addressed the problem with a single target that all three sensors can detect, and estimated the extrinsics from per-modality detections through a joint optimization that keeps the three sensor pairs consistent.
While effective in controlled settings, the specially designed target restricts the method to an offline setup.
We instead propose an online target-less joint calibration framework that constructs residuals for each sensor pair from natural features, and optimizes the extrinsics of all pairs together to minimize the overall residual.

\section{Joint radar-LiDAR-camera calibration}
\label{sec:methodology}

\begin{figure*}[t!]
    \captionsetup{font=footnotesize}
    \centering
    \includegraphics[width=\textwidth]{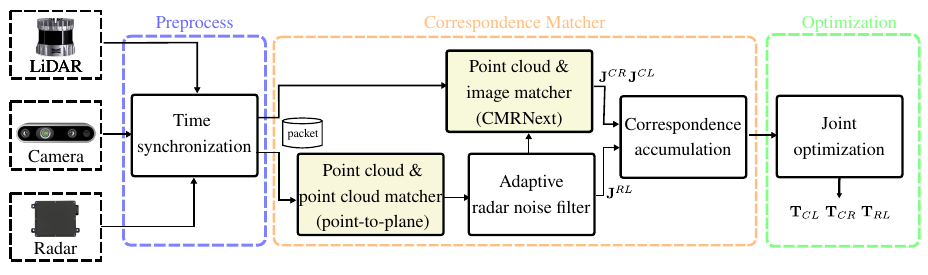}
    \caption{Overview of the proposed joint radar-LiDAR-camera calibration pipeline. Time-synchronized measurements are processed by two correspondence matchers: a point cloud and image matcher (CMRNext) for the camera-involving pairs and a point-to-plane matcher for the radar-LiDAR pair, where the adaptive radar noise filter rejects spurious radar correspondences. The correspondences are accumulated over frames, and the joint optimization estimates the extrinsics of all three sensor pairs.}
    \label{fig:overall_pipeline}
\end{figure*}

Our proposed radar-LiDAR-camera calibration pipeline consists of five modules: preprocessor, adaptive radar noise filter, correspondence matcher, correspondence accumulation, and joint optimization.
An overview of the proposed method is illustrated in Fig.~\ref{fig:overall_pipeline}.

\subsection{Preprocessor}
\label{subsec:preprocessor}
Preprocessor receives raw radar, LiDAR, and camera measurements asynchronously.
First, it gathers measurements from each sensor within a predefined time window based on their timestamps.
Since extrinsic calibration requires high-quality data rather than continuous streams, we set a tight time window of 0.03 seconds and discard measurements with large temporal gaps to minimize motion distortion. 
Then, synchronized measurements are transformed into a common coordinate frame using the initial extrinsic parameters and passed to the correspondence matcher.
\subsection{Adaptive Radar Noise Filter}
\label{subsec:adaptive_radar_noise_filter}

Radar measurements often contain spurious returns, mostly from the ground, which degrade calibration performance unless properly removed.
However, a fixed threshold cannot be reliably applied, since the radar's orientation relative to the ground is unknown prior to calibration.
To address this, we exploit the fact that the initial extrinsic parameters, although coarse, suffice to bound the region in which valid radar measurements should appear.

Each radar point $\dRadarPtInRadarFrame$ can be transformed into the LiDAR frame using the true extrinsics:
\begin{equation}
    \dRadarPtInLidarFrame = \dRotFromFrameLidarToRadar \, \dRadarPtInRadarFrame + \dTlsFromFrameLidarToRadar ,
    \label{eq:radar_to_lidar}
\end{equation}
where $\dRadarPtInLidarFrame$ and $\dRadarPtInRadarFrame$ are the radar point in the LiDAR and radar frames, respectively, and $\dRotFromFrameLidarToRadar$ and $\dTlsFromFrameLidarToRadar$ are the rotation and translation of the radar-to-LiDAR extrinsics.
Let $\dRotPertFromFrameLidarToRadar \in \mathbb{R}^{3}$ and $\dTlsPertFromFrameLidarToRadar \in \mathbb{R}^{3}$ be the initial extrinsic perturbation.
The nominal projection, available before calibration converges, applies the true extrinsics together with this perturbation through the $\dSO3$ exponential map $\dBigExp(\cdot)$:
\begin{equation}
    \dRadarPtNomInLidarFrame = \dRotFromFrameLidarToRadar \dBigExp\left( \dRotPertFromFrameLidarToRadar \right) \dRadarPtInRadarFrame + \dTlsFromFrameLidarToRadar + \dTlsPertFromFrameLidarToRadar .
    \label{eq:nominal_projection}
\end{equation}

For small $\dRotPertFromFrameLidarToRadar$, applying the first-order approximation $\dBigExp(\dRotPertFromFrameLidarToRadar) \approx \mathbf{I} + [\dRotPertFromFrameLidarToRadar]_{\times}$ to \eqref{eq:nominal_projection} and subtracting \eqref{eq:radar_to_lidar} yields the LiDAR-frame displacement induced by the unknown perturbation:
\begin{equation}
    \dRadarPtNoiseInLidarFrame
    \;:=\; \dRadarPtNomInLidarFrame - \dRadarPtInLidarFrame
    \;\approx\; -\dRotFromFrameLidarToRadar \, [\dRadarPtInRadarFrame]_{\times} \, \dRotPertFromFrameLidarToRadar
    \;+\; \dTlsPertFromFrameLidarToRadar ,
    \label{eq:projection_displacement}
\end{equation}
where $[\cdot]_{\times}$ denotes the skew-symmetric operator.
Taking norms of \eqref{eq:projection_displacement} with the triangle inequality, and assuming the initial perturbation is bounded by $\| \dRotPertFromFrameLidarToRadar \| \le \theta_{\max}$ and $\| \dTlsPertFromFrameLidarToRadar \| \le t_{\max}$, the displacement is bounded as
\begin{equation}
    \left\| \dRadarPtNoiseInLidarFrame \right\|
    \;\le\; \theta_{\max} \left\| \dRadarPtInRadarFrame \right\| + t_{\max} .
    \label{eq:displacement_bound}
\end{equation}
The admissible displacement grows linearly with the radar range $\|\dRadarPtInRadarFrame\|$---the same angular error $\theta_{\max}$ translates to a larger lateral error at long range---so a fixed threshold either rejects valid far points or accepts spurious near points, whereas the worst-case bound in \eqref{eq:displacement_bound} is adaptive by construction.

Combining \eqref{eq:displacement_bound} with the LiDAR measurement precision $\epsilon_{L}$, the maximum gap between a true surface and the nearest LiDAR return, we define the adaptive margin
\begin{equation}
    \tau\!\left( \dRadarPtInRadarFrame \right)
    \;=\; \theta_{\max} \left\| \dRadarPtInRadarFrame \right\| + t_{\max} + \epsilon_{L} .
    \label{eq:adaptive_margin}
\end{equation}
Let $\mathcal{P}^{L}$ denote the current LiDAR scan.
A radar point $\dRadarPtInRadarFrame$ is regarded as valid if and only if its nearest LiDAR neighbor ${}^{\dLidarFrame} \mathbf{p}_{\mathrm{n}} \in \mathcal{P}^{L}$ satisfies
\begin{equation}
    \left\| {}^{\dLidarFrame} \mathbf{p}_{\mathrm{n}} - \dRadarPtNomInLidarFrame \right\|
    \;\le\; \tau\!\left( \dRadarPtInRadarFrame \right) ,
    \label{eq:validity_condition}
\end{equation}
otherwise it is regarded as noise and discarded.

\subsection{Correspondence Matcher}
\label{subsec:correspondence_matcher}
In the correspondence matcher, we extract correspondences from each sensor pair, which are then used to form the cost functions for joint optimization.
First, we transform all radar and LiDAR points into the camera frame using the initial (perturbed) extrinsic estimate.

For the two camera-involving pairs (camera--LiDAR and camera--radar), we adopt CMRNext~\cite{cattaneo2025CMRNext}, an optical-flow-based deep network that takes the camera image together with the camera-frame-projected point cloud and predicts a flow field refining each projected pixel toward its true image location.
The resulting 2D--3D correspondence sets are
\begin{equation}
    \begin{aligned}
        \mathcal{C}^{\dCameraFrame\dRadarFrame}_{i} &= \big\{ \big(\dJthRadarPtNomInCameraFrame,\; {}^{\dCameraFrame} \mathbf{u}_{j}\big) \big\}_{j=1}^{N^{\dCameraFrame\dRadarFrame}_{i}}, \\
        \mathcal{C}^{\dCameraFrame\dLidarFrame}_{i} &= \big\{ \big(\dKthLidarPtNomInCameraFrame,\; {}^{\dCameraFrame} \mathbf{u}_{k}\big) \big\}_{k=1}^{N^{\dCameraFrame\dLidarFrame}_{i}},
    \end{aligned}
    \label{eq:correspondences_camera}
\end{equation}
where $\dJthRadarPtNomInCameraFrame, \dKthLidarPtNomInCameraFrame \in \mathbb{R}^{3}$ denote the $j$-th radar and $k$-th LiDAR points projected into the camera frame, ${}^{\dCameraFrame} \mathbf{u}_{j}, {}^{\dCameraFrame} \mathbf{u}_{k} \in \mathbb{R}^{2}$ are their matched 2D pixel coordinates on the image plane, and $N^{\dCameraFrame\dRadarFrame}_{i}, N^{\dCameraFrame\dLidarFrame}_{i}$ are the numbers of correspondences in each set at time $t_{i}$.
Although originally designed for camera--LiDAR, CMRNext transfers naturally to camera--radar thanks to the structural similarity between radar and LiDAR point clouds, and it has shown robust performance across diverse sensor configurations and robot platforms.

For the LiDAR--radar pair, we simply perform a nearest-neighbor search.
Each radar point is paired with its closest LiDAR return, yielding
\begin{equation}
    \mathcal{C}^{\dLidarFrame\dRadarFrame}_{i} = \big\{ \big(\dLthRadarPtNomInCameraFrame,\; \dLthLidarPtNomInCameraFrame\big) \big\}_{l=1}^{N^{\dLidarFrame\dRadarFrame}_{i}},
    \label{eq:correspondences_lr}
\end{equation}
where $\dLthRadarPtNomInCameraFrame, \dLthLidarPtNomInCameraFrame \in \mathbb{R}^{3}$ denote the $l$-th matched radar and LiDAR points in the camera frame, and $N^{\dLidarFrame\dRadarFrame}_{i}$ is the number of such correspondences at time $t_{i}$.

The correspondence matcher emits per-frame sets $\mathcal{C}^{\dCameraFrame\dRadarFrame}_{i}, \mathcal{C}^{\dCameraFrame\dLidarFrame}_{i}, \mathcal{C}^{\dLidarFrame\dRadarFrame}_{i}$.
Because radar is sparse, the radar-related sets are then handed to correspondence accumulation before joint optimization.

\subsection{Correspondence Accumulation}
\label{subsec:correspondence_accumulation}
Radar measurements are typically sparse, so the per-frame correspondence sets from radar-related pairs, $\mathcal{C}^{\dCameraFrame\dRadarFrame}_{i}$ and $\mathcal{C}^{\dLidarFrame\dRadarFrame}_{i}$, are far smaller than the camera--LiDAR set $\mathcal{C}^{\dCameraFrame\dLidarFrame}_{i}$.
This imbalance can cause the joint optimization to over-rely on the camera--LiDAR pair, yielding suboptimal calibration results.

Since the extrinsic parameters remain constant during calibration, correspondences extracted at different times can be pooled across frames without bias, except for those on moving objects, which RANSAC-based PnP with a 2\,pixel threshold discards before the optimization.
Over a window of $M = 10$ frames, we accumulate the radar-related correspondences as
\begin{equation}
    \bar{\mathcal{C}}^{\dCameraFrame\dRadarFrame} = \bigcup_{i=1}^{M} \mathcal{C}^{\dCameraFrame\dRadarFrame}_{i}, \qquad
    \bar{\mathcal{C}}^{\dLidarFrame\dRadarFrame} = \bigcup_{i=1}^{M} \mathcal{C}^{\dLidarFrame\dRadarFrame}_{i}.
    \label{eq:accumulation}
\end{equation}
The camera--LiDAR pair is already dense per frame, so no accumulation is needed; we use only the latest frame's set $\mathcal{C}^{\dCameraFrame\dLidarFrame}_{M}$.

The three correspondence sets passed to the joint optimization are $\bar{\mathcal{C}}^{\dCameraFrame\dRadarFrame}, \mathcal{C}^{\dCameraFrame\dLidarFrame}_{M}, \bar{\mathcal{C}}^{\dLidarFrame\dRadarFrame}$.

\subsection{Joint Optimization}
\label{subsec:joint_optimization}
Given the accumulated correspondences, we jointly optimize the extrinsic transformations relating the radar, LiDAR, and camera.
Let $\mathbf{T}_{\dCameraFrame\dRadarFrame}, \mathbf{T}_{\dCameraFrame\dLidarFrame}, \mathbf{T}_{\dLidarFrame\dRadarFrame} \in \dSE3$ denote the extrinsic transformations for the camera--radar, camera--LiDAR, and LiDAR--radar pairs, respectively, each represented as a $4 \times 4$ homogeneous matrix.

We define a residual for each sensor pair --- reprojection error for the camera-involving pairs and point-to-point distance for the LiDAR--radar pair.
For the camera-involving pairs, the reprojection errors are
\begin{equation}
    \begin{aligned}
        r^{\dCameraFrame\dRadarFrame}_{j} &= {}^{\dCameraFrame}\mathbf{u}_{j} - \pi\!\left( \dJthRadarPtInCameraFrame \right), \\
        r^{\dCameraFrame\dLidarFrame}_{k} &= {}^{\dCameraFrame}\mathbf{u}_{k} - \pi\!\left( \dKthLidarPtInCameraFrame \right),
    \end{aligned}
    \label{eq:residual_cam}
\end{equation}
where $\pi(\cdot)$ is the camera projection function mapping a 3D point to its 2D pixel coordinates, and the camera-frame points
$\dJthRadarPtInCameraFrame := \mathbf{T}_{\dCameraFrame\dRadarFrame}\, {}^{\dRadarFrame}\dJthRadarPt^{\mathrm{h}}$, $\dKthLidarPtInCameraFrame := \mathbf{T}_{\dCameraFrame\dLidarFrame}\, {}^{\dLidarFrame}\dKthLidarPt^{\mathrm{h}}$ are computed with the optimization variables $\mathbf{T}_{\dCameraFrame\dRadarFrame}, \mathbf{T}_{\dCameraFrame\dLidarFrame}$, where $(\cdot)^{\mathrm{h}}$ denotes homogeneous coordinates.
In contrast, the matcher's $\dJthRadarPtNomInCameraFrame, \dKthLidarPtNomInCameraFrame$ use the fixed initial estimate.

For the LiDAR--radar pair, the point-to-point distance is
\begin{equation}
    r^{\dLidarFrame\dRadarFrame}_{l} = {}^{\dLidarFrame}\dLthLidarPt - \mathbf{T}_{\dLidarFrame\dRadarFrame}\, {}^{\dRadarFrame}\dLthRadarPt^{\mathrm{h}}.
    \label{eq:residual_lr}
\end{equation}

The three sensor pairs form a closed loop: transporting a radar-frame point to the camera frame directly via $\mathbf{T}_{\dCameraFrame\dRadarFrame}$ must equal transporting it via $\mathbf{T}_{\dLidarFrame\dRadarFrame}$ and then $\mathbf{T}_{\dCameraFrame\dLidarFrame}$.
The extrinsic transformations therefore satisfy
\begin{equation}
    \mathbf{T}_{\dCameraFrame\dRadarFrame} = \mathbf{T}_{\dCameraFrame\dLidarFrame}\,\mathbf{T}_{\dLidarFrame\dRadarFrame}.
    \label{eq:loop_closure}
\end{equation}
Substituting $\mathbf{T}_{\dLidarFrame\dRadarFrame} = \mathbf{T}_{\dCameraFrame\dLidarFrame}^{-1}\,\mathbf{T}_{\dCameraFrame\dRadarFrame}$ into \eqref{eq:residual_lr} eliminates $\mathbf{T}_{\dLidarFrame\dRadarFrame}$ as an independent variable, giving
\begin{equation}
    r^{\dLidarFrame\dRadarFrame}_{l} = {}^{\dLidarFrame}\dLthLidarPt - \big(\mathbf{T}_{\dCameraFrame\dLidarFrame}^{-1}\,\mathbf{T}_{\dCameraFrame\dRadarFrame}\big)\, {}^{\dRadarFrame}\dLthRadarPt^{\mathrm{h}}.
    \label{eq:residual_lr_closed}
\end{equation}

We formulate the joint optimization problem as
\begin{equation}
    \begin{aligned}
        \min_{\mathbf{T}_{\dCameraFrame\dRadarFrame}, \mathbf{T}_{\dCameraFrame\dLidarFrame}} \Bigg\{ &\,
        w^{\dCameraFrame\dRadarFrame} \sum_{j=1}^{|\bar{\mathcal{C}}^{\dCameraFrame\dRadarFrame}|} \big\|r^{\dCameraFrame\dRadarFrame}_{j}\big\|^{2} \\
        & + w^{\dCameraFrame\dLidarFrame} \sum_{k=1}^{|\mathcal{C}^{\dCameraFrame\dLidarFrame}_{M}|} \big\|r^{\dCameraFrame\dLidarFrame}_{k}\big\|^{2} \\
        & + w^{\dLidarFrame\dRadarFrame} \sum_{l=1}^{|\bar{\mathcal{C}}^{\dLidarFrame\dRadarFrame}|} \big\|r^{\dLidarFrame\dRadarFrame}_{l}\big\|^{2} \Bigg\},
    \end{aligned}
    \label{eq:joint_objective}
\end{equation}
where $w^{\dCameraFrame\dRadarFrame} = 0.75, w^{\dCameraFrame\dLidarFrame} = 1.0, w^{\dLidarFrame\dRadarFrame} = 0.5$ are fixed weights assigned per sensor pair to balance their contributions, chosen empirically and kept fixed for every route.

Jointly optimizing across all sensor pairs lets reliable correspondences from one pair compensate for sparse or noisy measurements in another, yielding more robust and accurate extrinsic estimates.

\begin{table*}[t!]
    \captionsetup{font=footnotesize}
    \setlength{\tabcolsep}{3pt}
    \centering
    \caption{Quantitative comparison of the proposed method against the CMRNext baseline across three routes of our in-house dataset. $R$ [°] and $T$ [m] indicate rotation and translation errors. Best results per column are highlighted in bold.}
    \resizebox{\textwidth}{!}{
        \begin{tabular}{l|cccccc|cccccc|cccccc}
            \hline
            Sequence & \multicolumn{6}{c|}{Route 1} & \multicolumn{6}{c|}{Route 2} & \multicolumn{6}{c}{Route 3} \\
            \cline{2-19}
            Sensor pair & \multicolumn{2}{c}{R2C} & \multicolumn{2}{c}{L2C} & \multicolumn{2}{c|}{R2L} & \multicolumn{2}{c}{R2C} & \multicolumn{2}{c}{L2C} & \multicolumn{2}{c|}{R2L} & \multicolumn{2}{c}{R2C} & \multicolumn{2}{c}{L2C} & \multicolumn{2}{c}{R2L} \\
            \cline{2-19}
            Error metric & $R$ [°] & $T$ [m] & $R$ [°] & $T$ [m] & $R$ [°] & $T$ [m] & $R$ [°] & $T$ [m] & $R$ [°] & $T$ [m] & $R$ [°] & $T$ [m] & $R$ [°] & $T$ [m] & $R$ [°] & $T$ [m] & $R$ [°] & $T$ [m] \\
            \hline
            CMRNext & 2.56          & 0.47          & 0.62          & \textbf{0.34} & 2.71          & 0.63          & 1.39          & 0.53          & 0.45          & \textbf{0.36} & 1.42          & 0.68          & 2.06          & 0.56          & \textbf{0.74} & 0.36          & 2.05          & 0.68          \\
            Ours    & \textbf{0.64} & \textbf{0.41} & \textbf{0.56} & 0.36          & \textbf{0.47} & \textbf{0.23} & \textbf{0.50} & \textbf{0.34} & \textbf{0.40} & \textbf{0.36} & \textbf{0.31} & \textbf{0.15} & \textbf{0.88} & \textbf{0.36} & 0.76          & \textbf{0.34} & \textbf{0.64} & \textbf{0.14} \\
            \hline
        \end{tabular}
    }
    \label{tab:quantitative_results}
\end{table*}

\section{Experimental results}
\label{sec:experiments}

\subsection{Experiment Setup}
\label{subsec:experiment_setup}

We compare the proposed method against CMRNext~\cite{cattaneo2025CMRNext}, a state-of-the-art deep learning approach for camera-LiDAR extrinsic calibration.

For evaluation, we use an in-house dataset providing 4D imaging radar, LiDAR, and camera measurements collected across diverse urban environments, including parking lots and city streets.
The dataset is collected from a vehicle equipped with a Continental ARS 548 RDI 4D imaging radar, a Hesai AT128 LiDAR, and an RGB camera with a 120$^{\circ}$ horizontal field of view, driven at speeds between 5 and 21\,km/h.
Ground-truth extrinsics are obtained through offline target board-based calibration following~\cite{domhof2019Extrinsic}.
The training set consists of approximately 7{,}600 time-synchronized sensor triplets collected from an outdoor parking lot and an adjacent urban district.
For a fair comparison, we fine-tune CMRNext on this training set from the KITTI-pretrained weights, separately for LiDAR and for radar, and compose its radar-to-LiDAR result as $\mathbf{T}_{\dLidarFrame\dRadarFrame} = \mathbf{T}_{\dCameraFrame\dLidarFrame}^{-1}\mathbf{T}_{\dCameraFrame\dRadarFrame}$.
We evaluate on three test routes that are geographically disjoint from the training area and selected to cover distinct driving conditions: Route 1 traverses narrow alleys and streets crowded with dynamic objects such as pedestrians and vehicles, Route 2 covers typical urban streets also crowded with dynamic objects, and Route 3 consists of wide avenues with predominantly open space.
Each route comprises one minute of driving data.

To simulate calibration drift, we randomly perturb the ground-truth camera extrinsics from their initial values by up to 5° in rotation and 0.5\,m in translation, the bound $\theta_{\max}, t_{\max}$ of the adaptive radar noise filter, and we evaluate each method on 10 such perturbation samples per route.
For each sample, the method runs online over the entire route, and the reported estimate is the average of the per-frame extrinsic estimates produced during the operation.
The proposed method uses a single parameter set across all three routes, without per-route tuning.
We report the root mean square error~(RMSE) of rotation (in degrees) and translation (in meters), averaged over the 10 samples, with all methods evaluated under identical perturbations for a fair comparison.
The errors are reported separately for the three sensor pairs: radar-camera (R2C), LiDAR-camera (L2C), and radar-LiDAR (R2L).

\subsection{Quantitative Results}
\label{subsec:quantitative}

\begin{figure}[t!]
    \captionsetup{font=footnotesize}
    \centering
    \includegraphics[width=\linewidth]{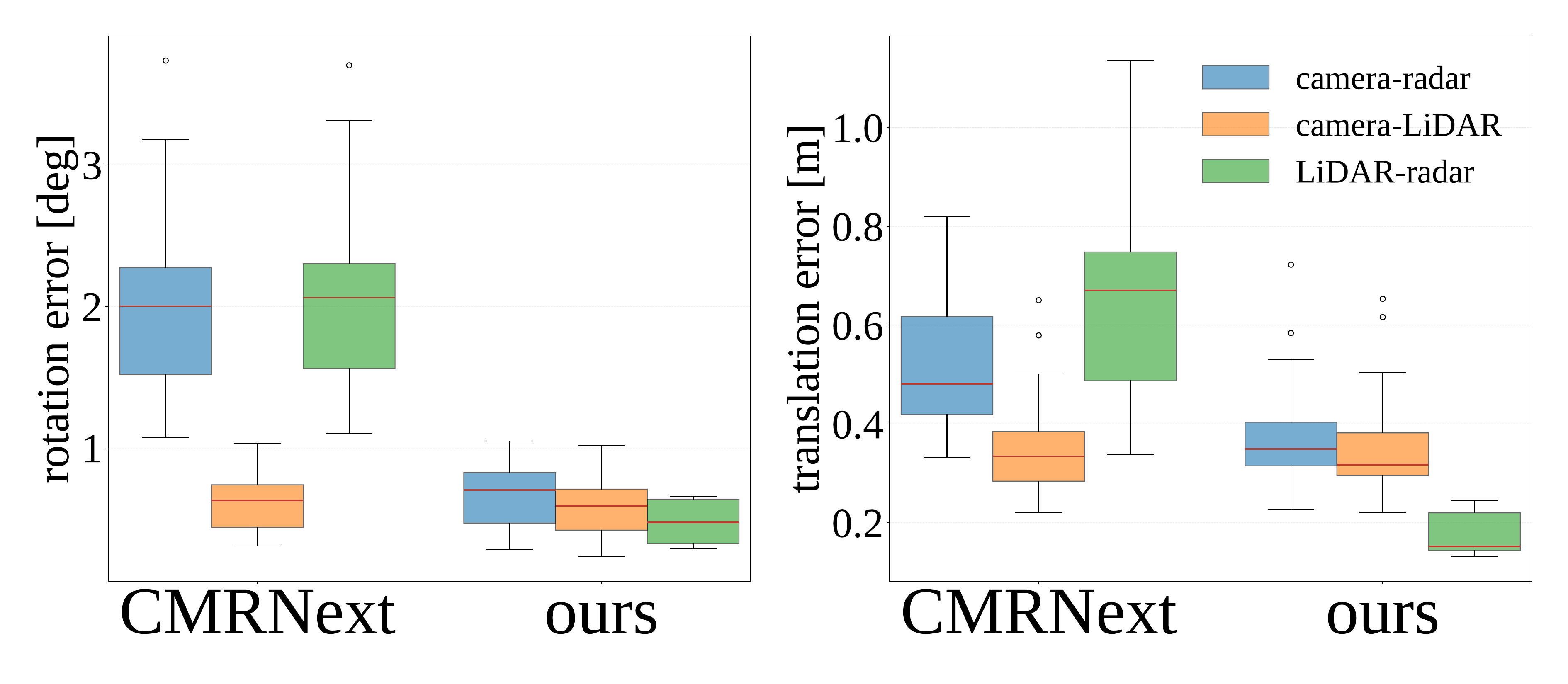}
    \caption{Distribution of rotation (left) and translation (right) errors of CMRNext and the proposed method over 10 random perturbation samples for the radar-camera (R2C), LiDAR-camera (L2C), and radar-LiDAR (R2L) sensor pairs. The proposed method achieves a lower median and a tighter spread across all three pairs.}
    \label{fig:boxplot_err}
\end{figure}

Table~\ref{tab:quantitative_results} compares the calibration accuracy of the proposed method and CMRNext across the three test routes, and Fig.~\ref{fig:boxplot_err} reports the corresponding error distributions over the 10 random perturbation samples.
The proposed method achieves the lowest error in most pair and metric combinations, with the largest gains on the radar-involving pairs (R2C and R2L).
Beyond the median, the boxplots also show a noticeably tighter spread for the proposed method, indicating more consistent calibration across perturbations.
For the R2C pair, the proposed method reduces the rotation error by an average of more than 60\% across the three routes, most on the crowded routes 1 and 2, and consistently lowers the translation error as well.
The R2L pair shows even larger gains, with rotation and translation errors reduced by roughly 75\% on average.
These improvements indicate that the joint optimization effectively compensates for the sparsity of radar measurements, which is the main bottleneck of pairwise radar calibrations.
For the L2C pair, where CMRNext already provides strong baseline performance, the proposed method maintains comparable or slightly better accuracy across all routes.
This shows that the joint formulation does not degrade the well-conditioned L2C pair while substantially improving the more challenging radar pairs, validating our motivation for jointly optimizing all sensor pairs.

\begin{table}[t!]
    \captionsetup{font=footnotesize}
    \setlength{\tabcolsep}{3pt}
    \centering
    \caption{Ablation study of the proposed components (ARNF, CA, JO) against the CMRNext baseline, averaged over all three test routes of our in-house dataset. $R$ [°] and $T$ [m] indicate rotation and translation errors. Best results per column are highlighted in bold.}
    \resizebox{\linewidth}{!}{
        \begin{tabular}{l|cc|cc|cc}
            \hline
            Sensor pair & \multicolumn{2}{c|}{R2C} & \multicolumn{2}{c|}{L2C} & \multicolumn{2}{c}{R2L} \\
            \cline{2-7}
            Error metric & $R$ [°] & $T$ [m] & $R$ [°] & $T$ [m] & $R$ [°] & $T$ [m] \\
            \hline
            CMRNext            & 2.00          & 0.52          & 0.60          & \textbf{0.35} & 2.06          & 0.66          \\
            Ours (ARNF)        & 1.71          & 0.46          & 0.60          & \textbf{0.35} & 1.79          & 0.57          \\
            Ours (ARNF+CA)     & 1.39          & 0.60          & \textbf{0.58} & 0.36          & 1.50          & 0.69          \\
            Ours (ARNF+CA+JO)  & \textbf{0.68} & \textbf{0.37} & \textbf{0.58} & \textbf{0.35} & \textbf{0.47} & \textbf{0.17} \\
            \hline
        \end{tabular}
    }
    \label{tab:ablation_study}
\end{table}

\subsection{Ablation Study}
\label{subsec:ablation}

To assess the contribution of each proposed component, we progressively add the adaptive radar noise filter (ARNF), correspondence accumulation (CA), and joint optimization (JO) modules to the CMRNext baseline.
Table~\ref{tab:ablation_study} reports the rotation and translation errors averaged over all three test routes.

Each added component contributes a measurable reduction in the rotation errors of both radar-involving pairs (R2C and R2L).
Stacking all three components reduces the rotation error by approximately 65\% on R2C and 77\% on R2L, while the translation errors are reduced by roughly 30\% and 74\%, respectively.
CA alone slightly increases the radar-pair translation errors, since the accumulated correspondences bias the purely pairwise optimization; this trade-off is recovered once JO is enabled.

In contrast, the L2C pair remains stable across all configurations, as CMRNext already provides accurate L2C correspondences.
This shows that the proposed components effectively recover accuracy on the more challenging radar pairs without degrading the well-conditioned L2C pair.

\begin{table}[t!]
    \captionsetup{font=footnotesize}
    \setlength{\tabcolsep}{2pt}
    \centering
    \caption{Runtime comparison of the proposed components against the CMRNext baseline. Configurations with CA accumulate correspondences over 10 frames before each calibration update; we therefore report both the per-update cost and the amortized per-frame cost.}
    \resizebox{\linewidth}{!}{
        \begin{tabular}{l|ccc}
            \hline
            Method            & \makecell{Accum. frames\\per update} & \makecell{Time per\\update [s]} & \makecell{Per-frame\\{[s]}} \\
            \hline
            CMRNext           &  1  & 0.67  & 0.67 \\
            Ours (ARNF)       &  1  & 0.68  & 0.68 \\
            Ours (ARNF+CA)    & 10  & 0.67  & 0.07 \\
            Ours (ARNF+CA+JO) & 10  & 10.64 & 1.06 \\
            \hline
        \end{tabular}
    }
    \label{tab:runtime}
\end{table}

\subsection{Runtime Evaluation}
\label{subsec:runtime}

Table~\ref{tab:runtime} reports the runtime of each ablation configuration.
Configurations without CA trigger a calibration update at every frame, while configurations with CA accumulate correspondences over 10 frames before triggering an update; we therefore report both the per-update cost and the amortized per-frame cost.

On a per-update basis, ARNF adds negligible overhead over the CMRNext baseline (0.68 vs.\ 0.67\,s), and CA alone keeps the per-update cost essentially unchanged (0.67\,s).
Joint optimization (JO), in contrast, raises the per-update cost to 10.64\,s due to the non-linear optimization over the cross-pair residuals.

However, CA amortizes the per-update cost of the full method over 10 frames, yielding a per-frame equivalent of only 1.06\,s.
This is roughly 1.6 times slower than the CMRNext baseline (0.67\,s/frame), even though JO is more than 15 times more expensive per update.
The combination of CA and JO therefore yields a favorable accuracy-runtime trade-off: CA absorbs the cost of the joint optimization while delivering the largest accuracy gain reported in \secref{subsec:ablation}.

\section{Conclusion}
\label{sec:conclusion}

In this paper, we presented an online target-less extrinsic calibration framework that jointly optimizes the extrinsics of all sensor pairs in the radar-LiDAR-camera system under a cross-pair constraint, along with an adaptive radar noise filter and a correspondence accumulation strategy for sparse and noisy radar measurements.
On an in-house dataset, the proposed method reduces the rotation errors of the radar-involving pairs by more than 60\% while maintaining comparable accuracy on the LiDAR-camera pair.
As future work, we plan to reduce the cost of the joint optimization and to add calibration monitoring that detects extrinsic drift.

\bibliographystyle{URL-IEEEtrans}
\bibliography{shin2026iccas_zotero_temp, URL-bib}

\end{document}